\documentclass{article}

     \usepackage[preprint]{neurips_2018}

\usepackage[utf8]{inputenc} 
\usepackage[T1]{fontenc}    
\usepackage[utf8]{inputenc}
\usepackage[export]{adjustbox}
\usepackage{mathtools}
\usepackage{hyperref}       
\usepackage{url}            
\usepackage{booktabs}       
\usepackage{amsfonts}       
\usepackage{nicefrac}       
\usepackage{microtype}      
\usepackage{graphicx}
\usepackage[font=small,labelfont=bf]{caption}
\usepackage{subcaption}
\usepackage{makecell}
\usepackage{wrapfig}
\usepackage{amsmath}
\usepackage{algorithm}
\usepackage[noend]{algpseudocode}
\usepackage{xcolor}
\usepackage{changepage}

\makeatletter
\def\BState{\State\hskip-\ALG@thistlm}
\makeatother

\title{Mass Estimation from Images using Deep Neural Network and Sparse Ground Truth}

\author{%
  Muhammad K.A.~Hamdan\thanks{Department of Electrical and Computer Engineering. Alternative email address: mhamdan.usa@gmail.com. Github: https://github.com/mhamdan91/Mass\_Flow\_Estimation} \\
  Iowa State University\\
  \texttt{mhamdan@iastate.edu} \\
   \And
   Diane T. Rover \\
   Iowa State University \\
   \texttt{drover@iastate.edu} \\ 
   \And
  Matthew J. Darr \\
   Iowa State University \\
   \texttt{darr@iastate.edu} \\
   \And     
   John Just \\
   Iowa State University \\
   \texttt{justjo@iastate.edu} \\
}
\begin{document}

\maketitle

\begin{abstract}
  Supervised learning is the workhorse for regression and classification tasks, but the standard approach presumes ground truth for every measurement. In real world applications, limitations due to expense or general in-feasibility due to the specific application are common. In the context of agriculture applications, yield monitoring is one such example where simple-physics based measurements such as volume or force-impact have been used to quantify mass flow, which incur error due to sensor calibration. By utilizing semi-supervised deep learning with gradient aggregation and a sequence of images, in this work we can accurately estimate a physical quantity (mass) with complex data structures and sparse ground truth. Using a vision system capturing images of a sugarcane elevator and running bamboo under controlled testing as a surrogate material to harvesting sugarcane, mass is accurately predicted from images by training a DNN using only final load weights. The DNN succeeds in capturing the complex density physics of random stacking of slender rods internally as part of the mass prediction model, and surpasses older volumetric-based methods for mass prediction. Furthermore, by incorporating knowledge about the system physics through the DNN architecture and penalty terms, improvements in prediction accuracy and stability, as well as faster learning are obtained. It is shown that the classic nonlinear regression optimization can be reformulated with an aggregation term with some independence assumptions to achieve this feat. Since the number of images for any given run are too large to fit on typical GPU vRAM, an implementation is shown that compensates for the limited memory but still achieve fast training times. The same approach presented herein could be applied to other applications like yield monitoring on grain combines or other harvesters using vision or other instrumentation.  
  
\end{abstract}

\section{Introduction}
\begin{wrapfigure}{R}{0.25\textwidth}
	\vspace{-\intextsep}
	\includegraphics[width=.25\textwidth]{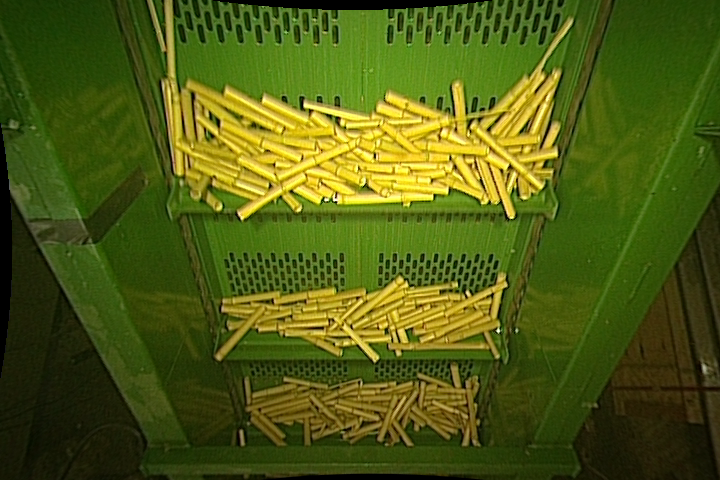}
	\caption{Sample image from the working dataset}
	\label{fig:dsample}
	\vspace{-4mm}
\end{wrapfigure}

Advances over the past couple of years in machine learning architectures and training methods, along with supporting open source frameworks such as Tensorflow, has made leveraging complex high-dimension datasets more accessible to scientists and engineers in various domains than ever before.  Along with this, various methods in the domain of machine learning have been developed to accommodate the variability, structure, and complexity of different research problems. Supervised learning has been most widely used but this method requires vast amount of labeled training data. The process of labeling data can be expensive, difficult, and time consuming when dealing with very large datasets. In many real world scenarios it is common not to have ground truth for every data point in the dataset, which poses a challenge to supervised learning. Unsupervised learning methods aiming to learn from data without labels have shown potential on simplistic data sets but have their drawbacks. In the presence of sufficient labeled data, unsupervised learning cannot achieve the accuracy levels of supervised learning, since there is no guarantee that the patterns the algorithm finds correlate directly with the ground truth or characteristics of interest \cite{lison2015introduction, marsland2011machine}. The work presented herein represents a middle-ground between the two aforementioned approaches, employing a version of semi-supervised learning by utilizing sparsely labeled data (i.e. data where not every measurement is labeled) for training algorithms. Semi-supervised learning has been leveraged in many applications such as pedestrian counting\cite{tan2011semi}, weed mapping in sunflower crops\cite{perez2015semi}, monocular depth map prediction~\cite{kuznietsov2017semi}, and object detectors from videos\cite{misra2015watch}. We leverage an agriculture application to prove the utility of this method by showing that predicting the mass from images of a complex material flowing on a sugarcane elevator, an otherwise difficult scenario to obtain ground truth for each image, is both rendered feasible and can replace older methods which use 3d point cloud information to estimate volume and subsequently convert to mass by (sometimes frequent) calibration while obtaining better performance. Some existing work to measure mass flow in similar context employs the use of LiDAR sensors \cite{jadhav2014volumetric}, and spectral analysis\cite{shin2012mass}. Accurate prediction of mass flow by images in this context enables precision agriculture through optimizing machine productivity and efficiency, and mitigating risk (e.g. by avoiding overfilling trucks) while making a more accessible low-cost solution since it only requires 2d camera and inference via a relatively light-weight neural network. The image in Figure~\ref{fig:dsample} is from the stereo camera mounted on the elevator and used in this work for volume estimation, and mass estimation by deep learning directly on the image. Mass estimation is of great importance to several industries and can be quite challenging obtain robust and accurate predictions depending on the material and application. It can be a critical parameter to industries like gas and oil and can be an optimizing factor for industries like mineral processing \cite{vayrynen2013mass}.The presented methods are by no means limited to image-based applications or agriculture industry and are readily transferable to other mass flow sensing applications from grain harvesters to mining, can work with other data sources besides images, and can be extended to any application where it is possible and desirable to utilize sparse ground truth under some basic assumptions like those presented here.

In this work, we propose an approach to learn complex physics relationships between the bulk density, quantity, and location of material in images to accurately estimate mass, with only sparse ground truth. Develop a regularization term that enforces temporal smoothness to achieve more stable prediction signals and better accuracy on data-point level. Propose a training procedure that handles gradient computation in batch mode, since the number of images for any given run is too large to fit in vRAM on a typical GPU. Develop and implement a lightweight DNN architecture that can run efficiently on low performance and memory bound devices. The proposed mass flow estimation method is shown to outperform previous methods that estimate volume and hinge on a calibrated density to transform the volume estimate to a mass estimate, while being less sensitive to light and offering a significantly more cost-effective solution.
\vspace{-1mm}
\section{Background}
\subsection{Modified Nonlinear Regression Loss for Sparse Ground Truth}
Self-training, mixture models, co-training, multiview learning, and 3S-vector machines are common semi-supervised learning techniques\cite{zhu2009introduction}. Generally, these techniques mostly rely on input data when correlating with the expected response. In our proposed method we focus on the response itself, i.e. to take advantage of measurements with sparse ground truth to train a predictive model, a tractable relationship must be asserted and hold between the measurements and labels. As long as that relationship can be formed into a loss function that can then be optimized, a model can be trained. Reinforcement learning with a Markov process assumption is one such example of sparse rewards and a tractable relationship where methods such as policy gradients\cite{sutton2000policy} can be employed. In a simpler case like this one, measurements can all be treated independently, and when predictions from these measurements are aggregated with a summation to produce a total (accumulated) mass, a loss function for the predictive model can be realized. Thus, the typical nonlinear regression formulation using mean squared error (MSE) loss can be slightly modified to incorporate an additional aggregation term over the predictions for each run in $k$ runs as shown below in Equation~\ref{eq:1} .The only difference being that the predictions of image "$x_j$" over a given run are summed and then compared to the ground truth and scaled by length "n" of the run. This type of simple modification to nonlinear regression opens up a major opportunity to bypass more costly or infeasible situations to obtain labels or ground truth in order to train a predictive model. The application presented in this work is one such situation where it is much easier to measure an accumulated mass as opposed to trying to obtain an individual mass for each measurement, which would be highly infeasible in any practical sense.\begin{equation}
\label{eq:1}
L(y;\hat{y}) =\sum_{i=1}^{k}\frac{1}{n}\bigg(y_i - \hat{y}_i\bigg)^2 
\end{equation}
\vspace{-6mm}
\subsection{Data Summary}
\begin{figure}[!ht]
	\centering
	\vspace{-\intextsep}
	\includegraphics[width=.65\textwidth]{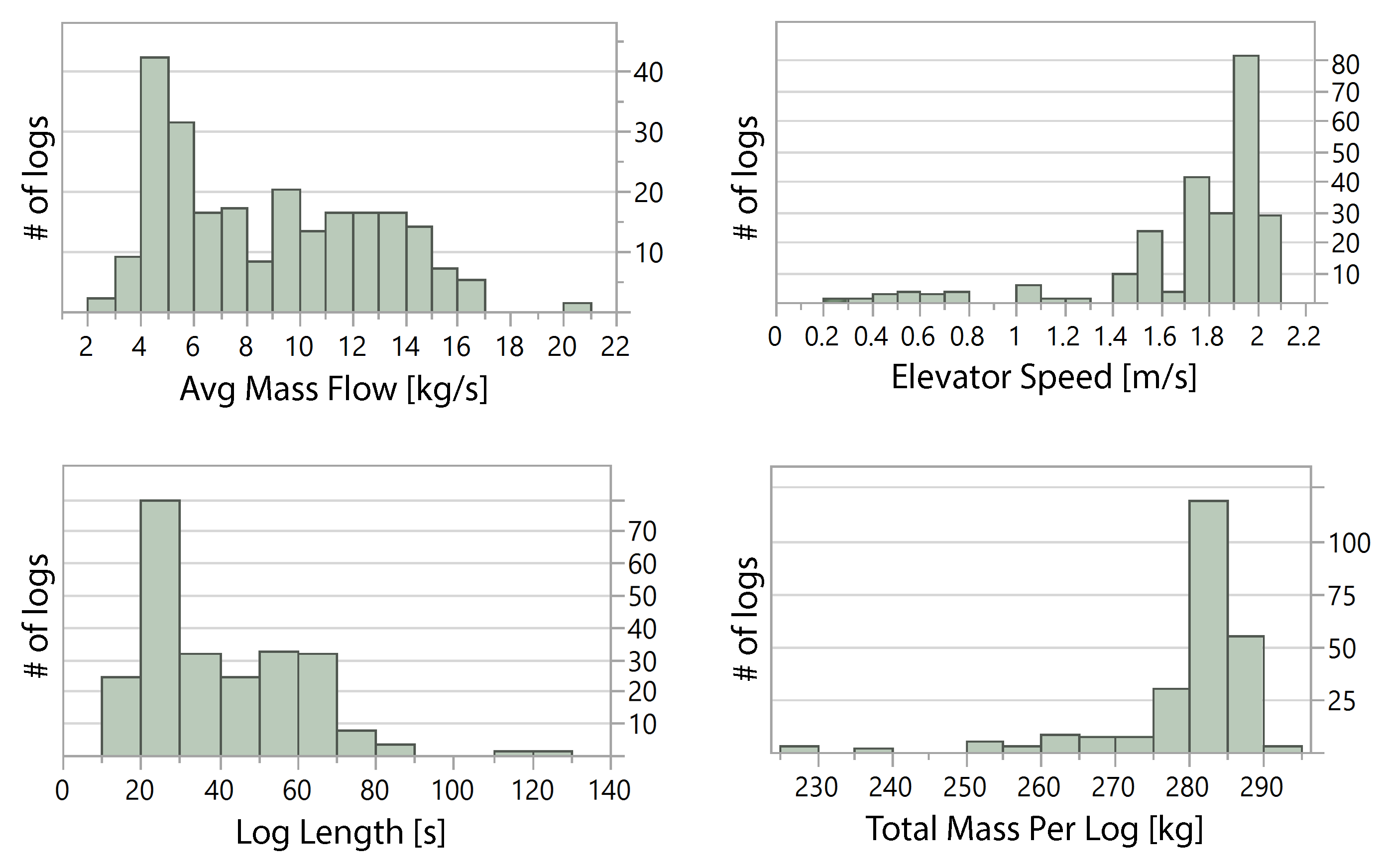} 
	\caption{Summary of data from testing runs showing range of factors covered - empty runs are not included.}
	\label{fig:dataset_description}
	\vspace{-6mm}
\end{figure}

To support development of a mass flow sensor based on stereo-camera volume measurements in previous unpublished work of our own, a design of experiment was conducted that took into account various factors that were projected to impact the response of the measurements and simulate environmental variables of operating conditions in a commercial application. These included factors such as variable illumination, mass flows, and elevator speeds. Six lighting levels were used that span from <0.7k lux up to $\sim$ 6.7k lux as measured by a light meter located on the elevator in front of the camera. Other factors are summarized by histograms of the average values per run as shown in Figure~\ref{fig:dataset_description} . Although some variation exists in the total mass of material (bamboo) used in each test run, this factor was held more consistent to focus on observing the impact of changing the other factors. A total of 239 runs were split into 60/20/20 - train/validation/test sets, which includes 8 empty runs (zero mass) running for $\sim$ 60 seconds with the elevator moving. The data from these tests is utilized for the 2d vision-based solution and compared to the results from volumetric-based mass prediction.

\subsection{Mass Flow by Volume}
\begin{figure}[!ht]
	\centering
	\vspace{-\intextsep}
	\begin{subfigure}[b]{0.4\textwidth}
		\centering
		\includegraphics[width=\textwidth]{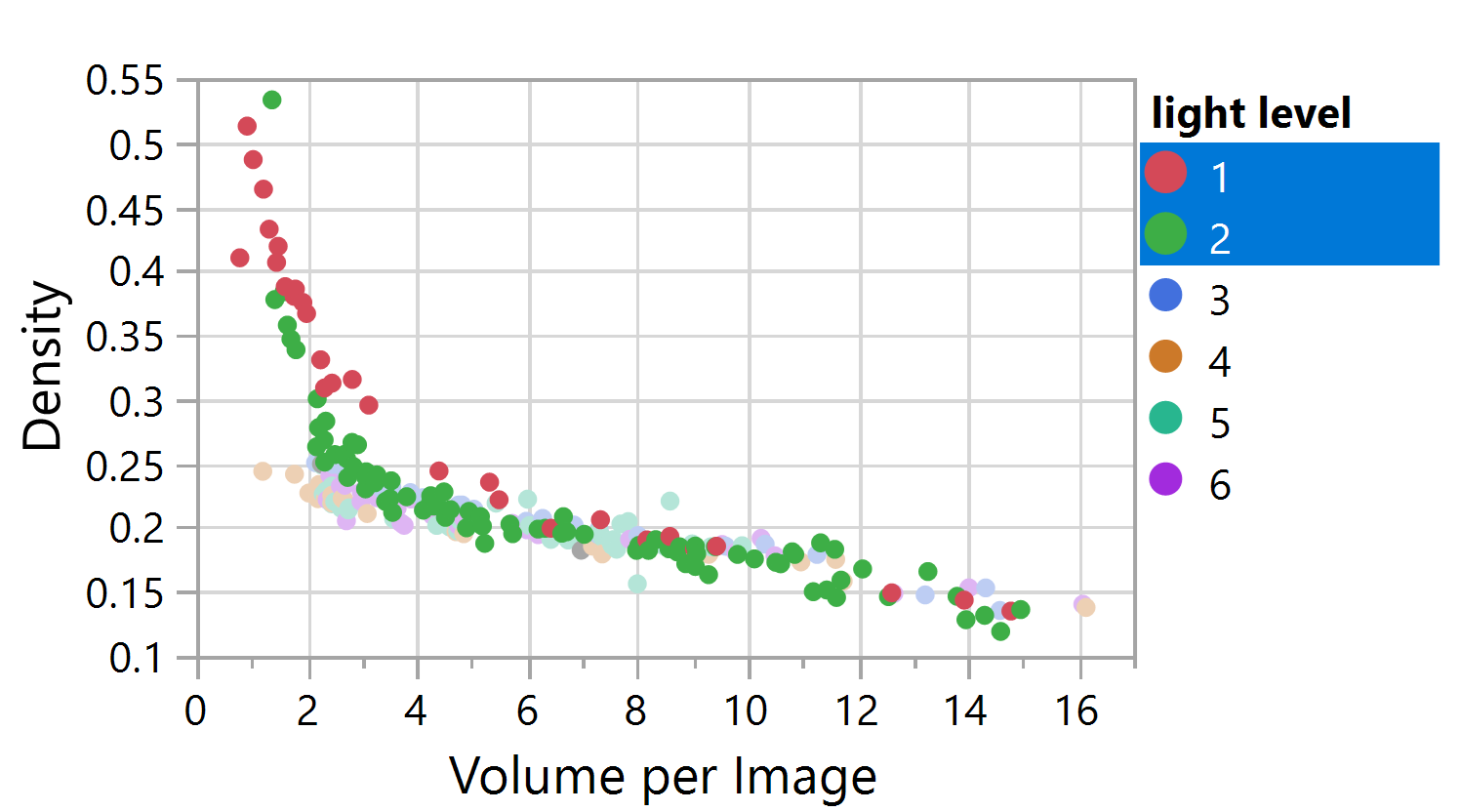}
		\caption{Lighting levels <2.7k lux}
		\label{fig:volume_plot_a}
	\end{subfigure}
	\hfill
	\begin{subfigure}[b]{0.4\textwidth}
		\centering
		\includegraphics[width=\textwidth]{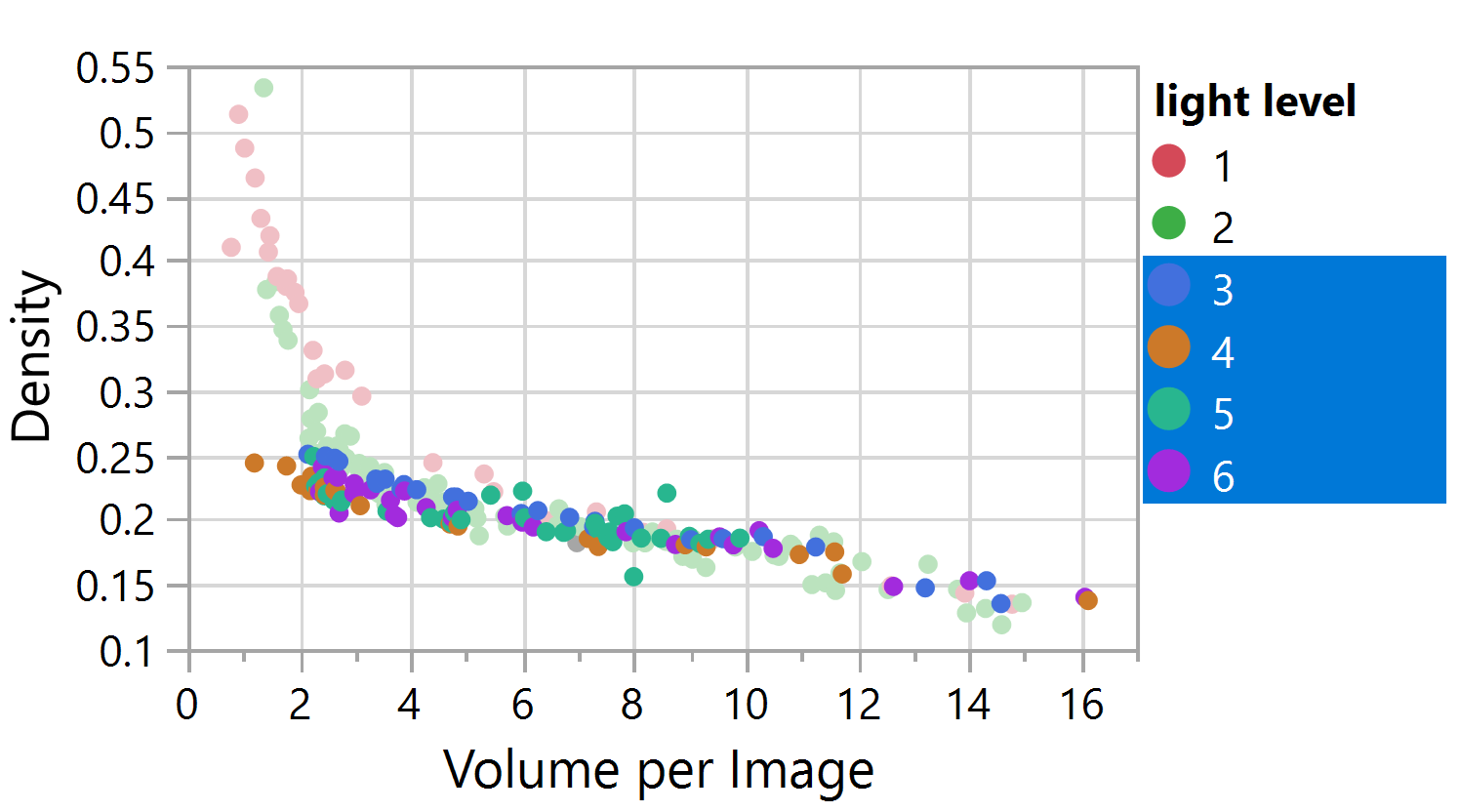}
		\caption{Lighting levels >2.7k lux}
		\label{fig:volume_plot_b}
	\end{subfigure}
	\caption{Left: Density dependence trends. Lighting levels <2.7k lux indicate poor sensitivity of stereo camera at lower volume flows. Right: runs with adequate lighting show decreasing bulk density with volume likely due to disordered stacking of material – a real/physical phenomenon}
	\label{fig:volume_plots}
	\vspace{-6mm}
\end{figure}

The baseline comparison for our implementation is a volume estimate obtained from a stereo camera at 7.5Hz. Previous work in sugar beet yield monitoring has used similar concepts with different instruments to obtain a volume measurement from which mass flow can be estimated \cite{konstantinovic2007soil}. The stereo camera produced a point cloud referenced to the plane of the elevator. Volume was calculated from this by binning the point cloud into squares in the $<x, y>$ plane and taking the median per bin to remain insensitive to outliers. Bins with a low number of points or standard deviation above a specific threshold were deemed unreliable and excluded from the volume estimation. Note that the calculated volume units from the stereo camera were scaled for convenience, and so the units used in this analysis are not labeled with physical units (nor do they matter for this application since it must be calibrated anyways). Likewise, the density is the total mass of the material divided by the total volume as calculated from the point cloud, and is shown as a unit-less number in the plots shown in Figure ~\ref{fig:volume_plots} . It was found during testing that, so long as lighting was kept at a certain minimum level (in this case > 2.7k Lux), significant nonlinearities with the point cloud estimation were avoided. These nonlinearities in low lighting manifested as increased number of points with low elevation (or “z”) values in each bin which dragged down the volume estimates. Since the shutter speed had to be fast due to the quick movement of the slats (and therefore material), increasing the exposure time was not an option. However as long as certain minimum amount of lighting is provided, a decreasing density still persists but is not as extreme. This remaining decreasing density trend is hypothesized due to a packing effect where the bulk density of the material decreases as the total volume of material increases. Work investigating the densities of disordered packing of slender cylinders supports this \cite{liu2018evolutions, zhang2006experimental}. Drawing inspiration from the trends in the above figures showing density decreasing with higher volumes, an algorithm was devised and fit that explicitly uses the stereo volume to predict the density, and then use it as a multiplier on the volume as is shown in Equation~\ref{eq:2} . This has some similarities in concept to an attention mechanism used in NLP semantic analysis\cite{bahdanau2014neural}, but in this case the weighted multiplier (density) for a given volume estimate is unbounded.
\begin{equation}
	\label{eq:2}
	Mass = f(max(V-\beta,0);\theta)\times max(V-\beta,0) \times v_{elev} \times t
\end{equation}
Where "f" is a 2-layer neural network parameterized by “$\theta$” that outputs a prediction of density based on the volume (V), scaled by elevator speed ($v_{elev}$) and capture time ($t$), with tanh activation and 3-hidden units. The $\beta$-parameter is also fit to account for any positive bias present since the stereo volume calculation was designed towards not missing any volume. This was also a meaningful formulation since if the $\beta$ parameter turned out to be negative then it would indicate a volume estimation in need of refinement, since it would be compensating for volume not detected. This was fit using the same aggregation technique as described in Equation ~\ref{eq:1} using a 2nd order optimizer

\section{Deep Learning with Sparse Ground Truth}
\subsection{Learning Mass from Images}
Learning mass from images is possible in this problem context due to the constraints imposed on the system and the ability to learn complex patterns via end-to-end training with deep learning. More specifically, the camera has a fixed sampling frequency that is fast enough to measure all the material passing by, it is mounted at a fixed distance from the elevator, and the only additional factor needed is slat velocity to scale the accumulated mass. Intuitively, the velocity is scaling the mass to produce a mass flow, since if the mass was sitting still it would not be accumulated. Alternatively it can be thought of as a way to account for frame overlap. Even though complex nonlinearities exist with the density and the material location in the image (e.g. the same material further away is smaller in the image), these patterns can be learned by optimizing a DNN due to fixed locations between the camera and the elevator. Understanding this reasoning about the system is crucial to designing and training a successful model in this and similar applications, since (e.g.) naively assuming counting pixels with material will correlate well with mass would lead to wasted efforts considering only a single factor such as image deformation. The value of using images of volumetric methods will potentially be enhanced further when observing harvested materials of sugarcane or grains, since in the case of sugarcane the constituents such as billets and leaves (and therefore density) vary considerably more . The constituents and their sizes are observable in images though and therefore patterns more than likely learnable by the same sparse ground truth methods. In grains the sizes of the kernels can change with moisture, among other things, and may exhibit other visual signs that correlate with density that an image-based algorithm could identify and learn to account for.
\subsection{Temporal Smoothing}
Utilizing prior knowledge about the problem allows more custom formulation of the model, loss function, and training procedure to improve accuracy and stability. In this application, it is clear that there is temporal correlation and images near in time should have more similarity in mass than images further away in time. To account for this during training, a regularizing term was added to the cost function with an associated hyper-parameter that allows penalizing a 1st-order lagged difference in the predicted mass values. Equation ~\ref{eq:3} shows full loss function for a run\textsubscript{i} that includes prediction error and the additional temporal smoothing regularization term in red.
\vspace{-2mm}
\begin{equation}
\label{eq:3}
L_i(x,y;w)=\frac{1}{n_i}\big\{y_i-\sum_{j=1}^{n_i}(f(x_{ij};w)\times v_{ij}\times t)\big\}^2 + \color{red}{ \frac{\lambda}{n_i}\sum_{j=1}^{n_i}\big\{f(x_{ij};w)-f(x_{i{(j-1)}};w)\big\}^2} \color{black} 
\vspace{-2mm}
\end{equation}
Prediction is corrected by the elevator speed $v$ and capture time $t$ (constant $7.5Hz$) to account of frame overlap and the loss is normalized by the number of images “n” in each run to equally weight the gradient update from each run. The penalty strength is controlled by a hyper-parameter $\lambda$ (chosen empirically $0.05$). To see this, note that the gradient of the loss function has a sum of gradients in it, which if left un-normalized will amount to larger gradient updates for longer runs (runs with more images) even though the runs mostly contain the same total mass.
\vspace{-1.6mm}
\subsection{Gradient Update}
To perform a gradient update, we need to compare our prediction, $\hat{y}_i$ with ground truth $y_i$, but our prediction and regularization terms contain sums, which means the gradient of the loss in Equation ~\ref{eq:3} does as well, as can be seen by Equation ~\ref{eq:4} . Since the number of images is too large to fit on a single GPU (a problem for very sparse ground truth), we keep a running sum of the gradients, predictions, and regularizing term from each batch to greatly reduce memory requirements. The full gradient can then be calculated and applied at the end of the run using the accumulated terms from the gradient along with the ground truth, run length, and regularizing hyper-parameter. Equation ~\ref{eq:4} describes gradient of the full form of the loss equation associated with one run or run “i”. Terms in red are accumulated from each batch of a run and the full gradient calculated and applied when the end of a run is reached during the training loop.
\begin{equation}
\label{eq:4}
\frac{\partial{L_i}}{\partial{w}} \gets -\frac{2}{n_i}\big[y_i - \color{red}\sum_{j=1}^{n_i}\hat{y}_{ij} \color{black}\big] \times \color{red} \sum_{j=1}^{n_i}\frac{\partial{\hat{y}_{ij}}}{\partial{w}} \color{black} + \frac{2\lambda}{n_i} \color{red}\sum_{j=1}^{n_i}{\bigg\{\big[\breve{y}_{ij}-\breve{y}_{i(j-1)}\big] \times \big[\frac{\partial{\breve{y}_{ij}}}{\partial{w}}-\frac{\partial{\breve{y}_{i(j-1)}}}{\partial{w}}\big]\bigg\}}\color{black}
\end{equation}
Given the large and variable size of runs, we compute gradients and predictions in batches and accumulate them over the course of their respective runs as shown in the Pseudo-code in algorithm~\ref{alg:the_alg} .
\setlength{\textfloatsep}{0.1cm}
\begin{algorithm}
	\caption{Computing and applying gradients over a single epoch. With each run\textsubscript{i} we have run length, images and speeds at the time of each image captured, and total ground truth mass}
	\label{alg:the_alg}
	\begin{algorithmic}[1]
		\For{\texttt{run} \textbf{in} \texttt{train\_data}}
		\For{$\texttt{batch, (images, speeds)} \hspace{0.1cm} \textbf{in}  \hspace{0.1cm}\texttt{run}\hspace{0.1cm}$}
		\State $x_b, v_b = \textbf{fetch\_next\_batch} \texttt{(images, speeds)}$\Comment{Data loaded in chronological order}
		\If{\texttt{New\_run}}
		\State $\texttt{run\_remainder} \gets \texttt{mod(run\_length, sizeof(batch))}$
		\State $\texttt{iterations} \gets \texttt{ceil(div(run\_length, sizeof(batch)))}$
		\State $\texttt{New\_run} \gets \texttt{False}$
		\EndIf
		\If {$\texttt{iterations > 1}$}	
		\State $\textit{\(\hat{y}_{b}$ $ \gets f(x_b;w)\)}$\Comment{Predict using DNN}
		\State $\textit{\(\hat{y}_{b_{vt}}\)} \gets \textit{\(\hat{y}_{b} \times v_b \times t\)}$\Comment{Correct frame-overlap}
		\State $\textit{\(\hat{y} \mathrel{{+}{=}} \Sigma \hat{y_b}_{vt},\hspace{0.5cm}  \hat{y_b}_{grad} \mathrel{{+}{=}}\Sigma \frac{\partial{ \hat{y_b}_{vt}}}{\partial{w}}\)}$\Comment{Accumulate predictions and gradients}
		\State $\textit{\(\hat{y}_{smooth} \mathrel{{+}{=}} \sum \{(\hat{y}_{b_{vt}}-\hat{y}_{(b_{vt}-1)})\)}$ $\odot$ $\textit{\((\frac{\partial \hat{y}_{b_{vt}}}{\partial w} - \frac{\partial \hat{y}_{(b_{vt}-1)}}{\partial w}) \}\)}$\Comment{Accumulate penalty}
		\State $\texttt{iterations \(\mathrel{{-}{=}} 1\)}$
		\Else
		\If {\texttt{batch} \textbf{contains} \texttt{curren\_run\_images\_ONLY}}
		\State$\textit{\(\frac{\partial L_{i}}{\partial w} \gets \frac{-2}{n_i}(y_{true} - \hat{y}) \odot \hat{y_b}_{grad}\)}$    $\textit{\(\oplus \frac{2 \lambda}{n_i} \hat{y}_{smooth} \)}$\Comment{Apply gradients}
		\State $ \textit{\(w \gets w + \alpha \frac{\partial L_{i}}{\partial w}\)}$\Comment{Update weights}
		\State $\texttt{New\_run} \gets \texttt{True}$
		\Else
		\State $\textbf{Do}$ $\texttt{steps 16 \(\to\) 18}$ $\textbf{for}$ $\texttt{images}$ $\textbf{in}$ $\texttt{current\_run}$
		\State $\textbf{Do}$ $\texttt{steps 3 \hspace{0.2cm}\(\to\) 13}$ $\textbf{for}$ $\texttt{images}$ $\textbf{in}$ $\texttt{next\_run}$
		\EndIf
		\EndIf
		\EndFor
		\EndFor
	\end{algorithmic}
\end{algorithm}
	\begin{figure}[!ht]
	\includegraphics[width=\textwidth]{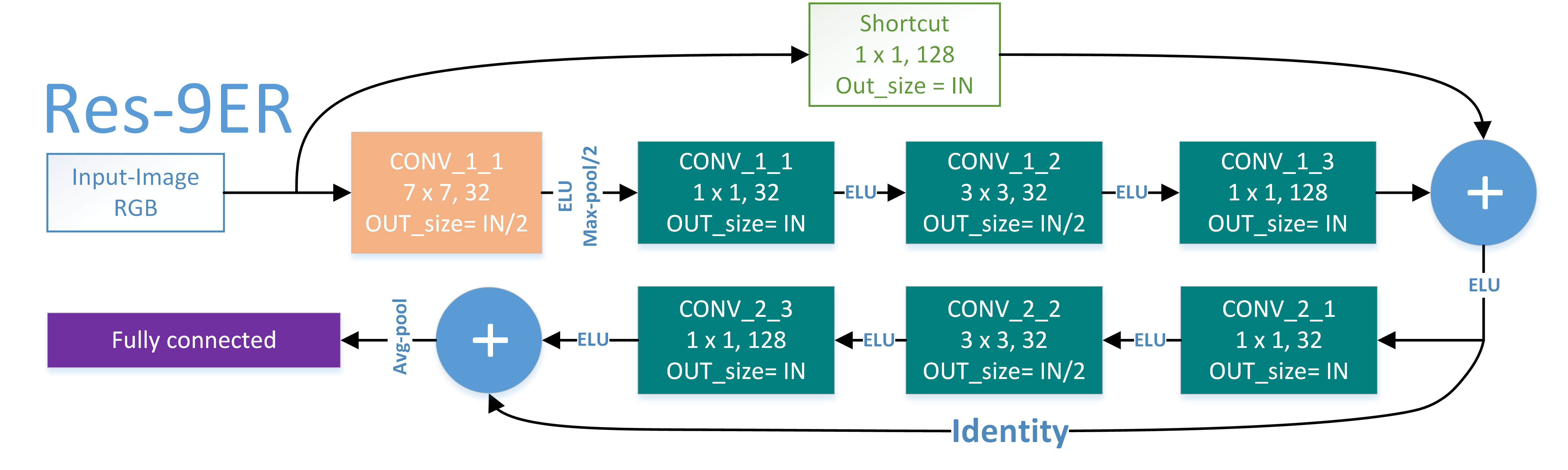} 
	\caption{Reduced residual 9 architecture (RES-9ER)}
	\label{fig:res_9er}
	\vspace{6mm}
\end{figure}
\subsection{Model Architecture and Training Procedure}
\begin{figure}[!ht]
	\centering
	\vspace{-7mm}
		\begin{subfigure}[t]{0.58\textwidth}
		\centering
		\includegraphics[width=\textwidth]{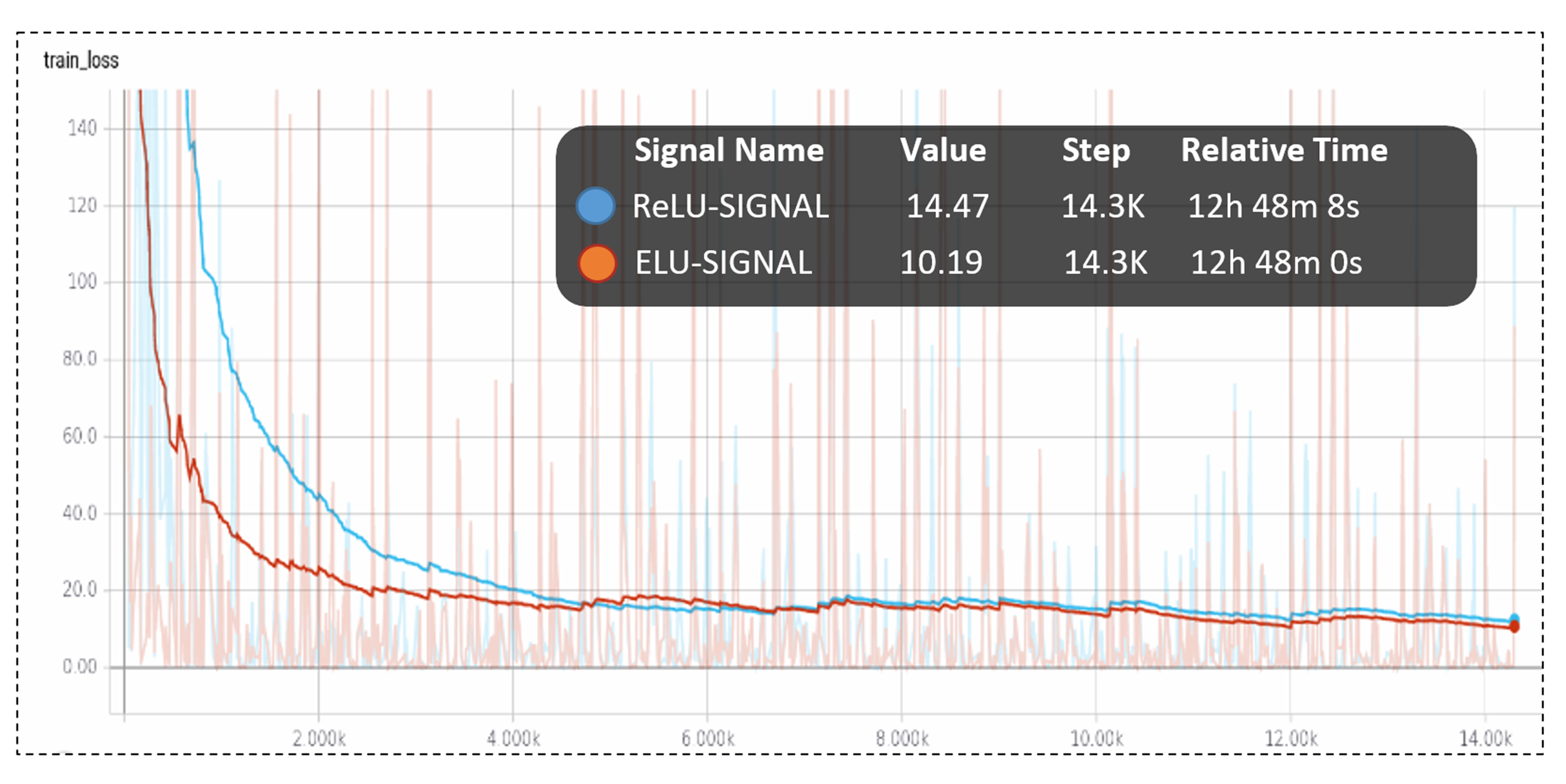}
		\caption{Training loss decay when using ReLU activation functions (in light blue) versus when using ELU activation functions (in orange). As shown, ELU loss signal converges faster than ReLU loss signal}
		\label{fig:loss_train}
	\end{subfigure}
	\hfill
	\begin{subfigure}[t]{0.38\textwidth}
		\centering
		\includegraphics[width=\textwidth]{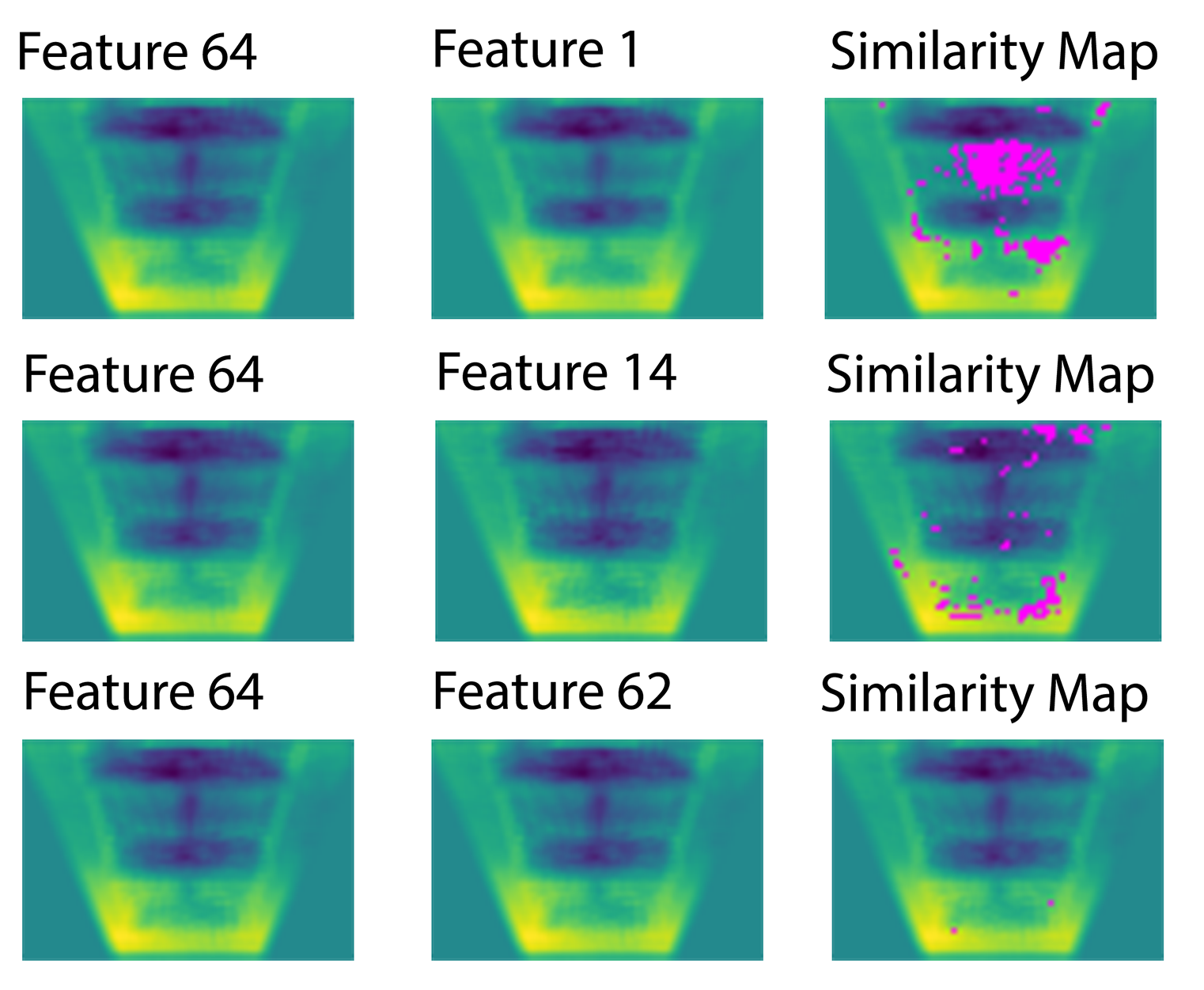}
		\caption{Left column shows reference feature, middle column shows resembling features and right column shows similarity maps between reference and resemblance features}
		\label{fig:f_maps}
	\end{subfigure}
	\caption{Training loss decay when using ReLU activation Vs ELU activation is shown in Fig.4a and redundant feature maps are shown in Fig.4b. Using these types of investigations helped inform of a suitable architecture that balanced accuracy, generalization, and stability.}
	\label{fig:loss_maps}
\end{figure}

\vspace{-3.5mm}
We consider complexity and size as essential factors in the developed architecture, where eventual application goes on embedded hardware at mass scale and saving every bit of computation to a minimum is highly desirable to save costs. To assess the complexity of our dataset we adopt a residual like style\cite{he2016identity} and follow a systematic approach in DNN design, starting shallow then reproducing accordingly. A 9-layer DNN with ELU units and residual connections, defined as – “RES9-ER” shown in Figure ~\ref{fig:res_9er}, was found to have good predictive accuracy as well as fast training and inference. Similar models with ReLU activation and 16 layers were also investigated before converging on this final architecture. ELU activation were considered in our DNN because it showed better noise dampening and more stable signal as well as helped converge faster than ReLUs as shown in Figure ~\ref{fig:loss_train} . The 16-layer network – “RES-16E”, barely out-performed the 9-layer network – “ RES-9E”. For this reason, we investigated the \begin{wraptable}{r}{5cm}
	\vspace{-2mm}
	\caption{Summary of developed architectures and their sizes}
	\label{tabl:networks_summary}
	\vspace{-4mm}
	\begin{center}
		\begin{tabular}{|p{1.2cm}|p{1.4cm}|p{1.2cm}|}\hline	
			\textbf{\scriptsize Architecture}      & \scriptsize\textbf{Number of parameters}     & \scriptsize\textbf{Number of layers} \\
			\cline{2-3}
			\hline
			\textbf{\scriptsize Res-16E}   & \texttt{\scriptsize 959489}    & \texttt{\scriptsize 16} \\
			\textbf{\scriptsize Res-9E}    & \texttt{\scriptsize 154113}    & \texttt{\scriptsize 9} \\
			\textbf{\scriptsize Res-9ER}   &  \texttt{\scriptsize 45921}    & \texttt{\scriptsize 9} \\
			\hline
		\end{tabular}
		\vspace{-6mm}
	\end{center}
\end{wraptable}learned features in every layer of Res-9E and observed redundant features, hence we reduced the number of filters in Res-9E to introduce reduced Res-9, the best performing architecture defined as – “RES-9ER”. Our observation of the redundant features is empirical and validated using this work \cite{resemblejs}.Figure ~\ref{fig:f_maps} shows a selected feature {64} against a set of other similar features {1, 14, 62}, whereas regions of different pixel information are highlighted in pink in the similarity map.Table~\ref{tabl:networks_summary} shows the sizes of RES-9E, RES-16, and RES-9ER.

\section{Results \& Analysis}
\subsection{Visualization Explanation of DNN Functioning}
\begin{figure}[!ht]
	\centering
	\vspace{-2mm}
	\begin{subfigure}[b]{0.3\textwidth}
		\centering
		\includegraphics[width=\textwidth]{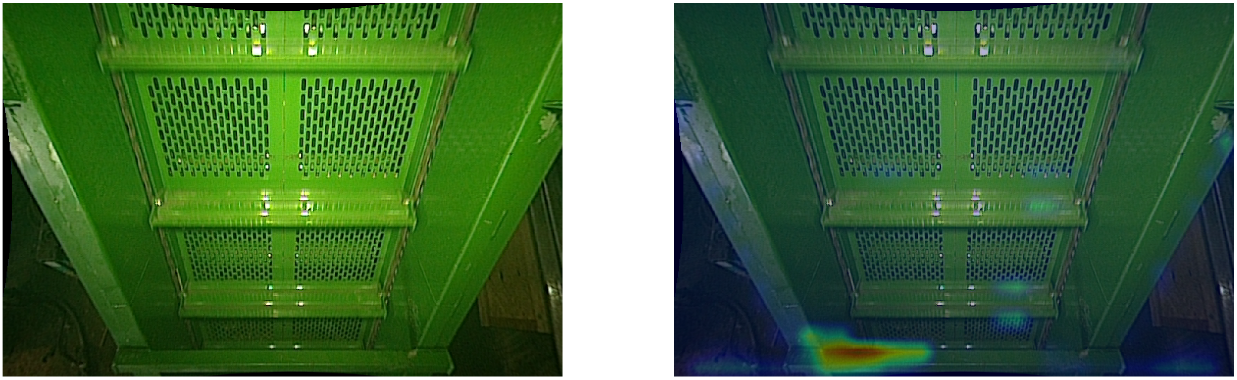}
		\caption{No flow of bamboo}
		\label{fig:empty_mass}
	\end{subfigure}
	\hfill
	\begin{subfigure}[b]{0.3\textwidth}
		\centering
		\includegraphics[width=\textwidth]{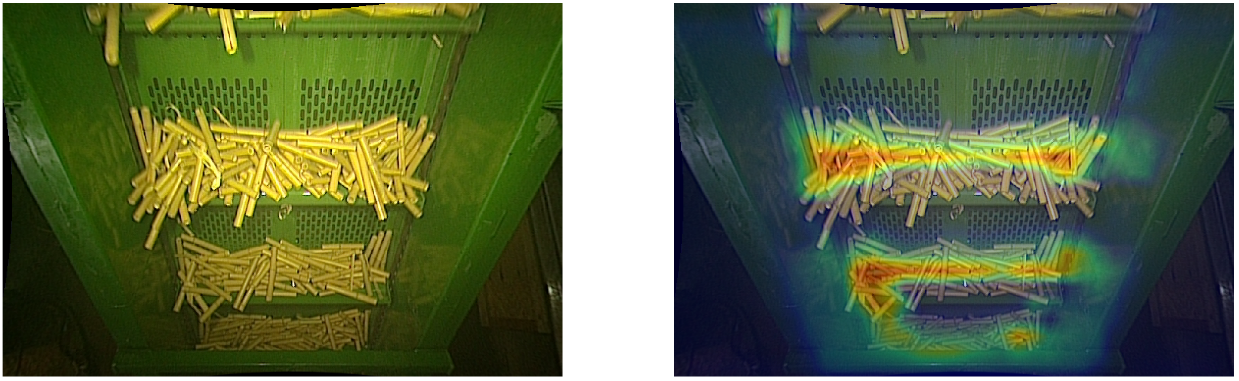}
		\caption{Moderate flow of bamboo}
		\label{fig:low_flow}
	\end{subfigure}
	\hfill
	\begin{subfigure}[b]{0.3\textwidth}
		\centering
		\includegraphics[width=\textwidth]{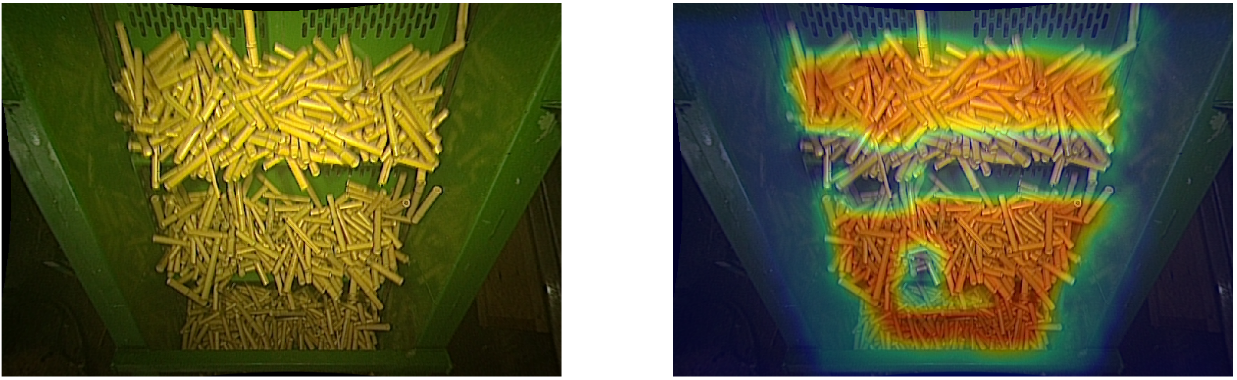}
		\caption{High flow of bamboo}
		\label{fig:high_flow}
	\end{subfigure}
	\caption{Images taken from different runs shown with their corresponding grad-cam++ heat-maps}
\label{fig:gradcam}
\vspace{-6mm}
\end{figure}

To get an intuition on what the network actually learns and what parts of the image contribute the most to the outcome, we follow a generalized gradient-based CNN visualization approach to visualize the learned mass in images. We adapt the grad-cam method proposed by this work \cite{gradcam2017} and develop an implementation that is compatible with eager mode in Tensorflow \cite{eager}. Shown in Figure ~\ref{fig:gradcam} , three different images and their corresponding grad-cam heat-maps.  As shown in Figure ~\ref{fig:empty_mass}, the network learns when there is not bamboo material present in the image as well as the network learns when there is various amounts of bamboo in the image as show by the grad-cam in Figure ~\ref{fig:low_flow}~\ref{fig:high_flow}.

\subsection{Temporal Smoothness Effect}
\begin{figure}[!ht]
	\centering
	\vspace{-2mm}
	\begin{subfigure}[b]{0.325\textwidth}
		\centering
		\includegraphics[width=\textwidth]{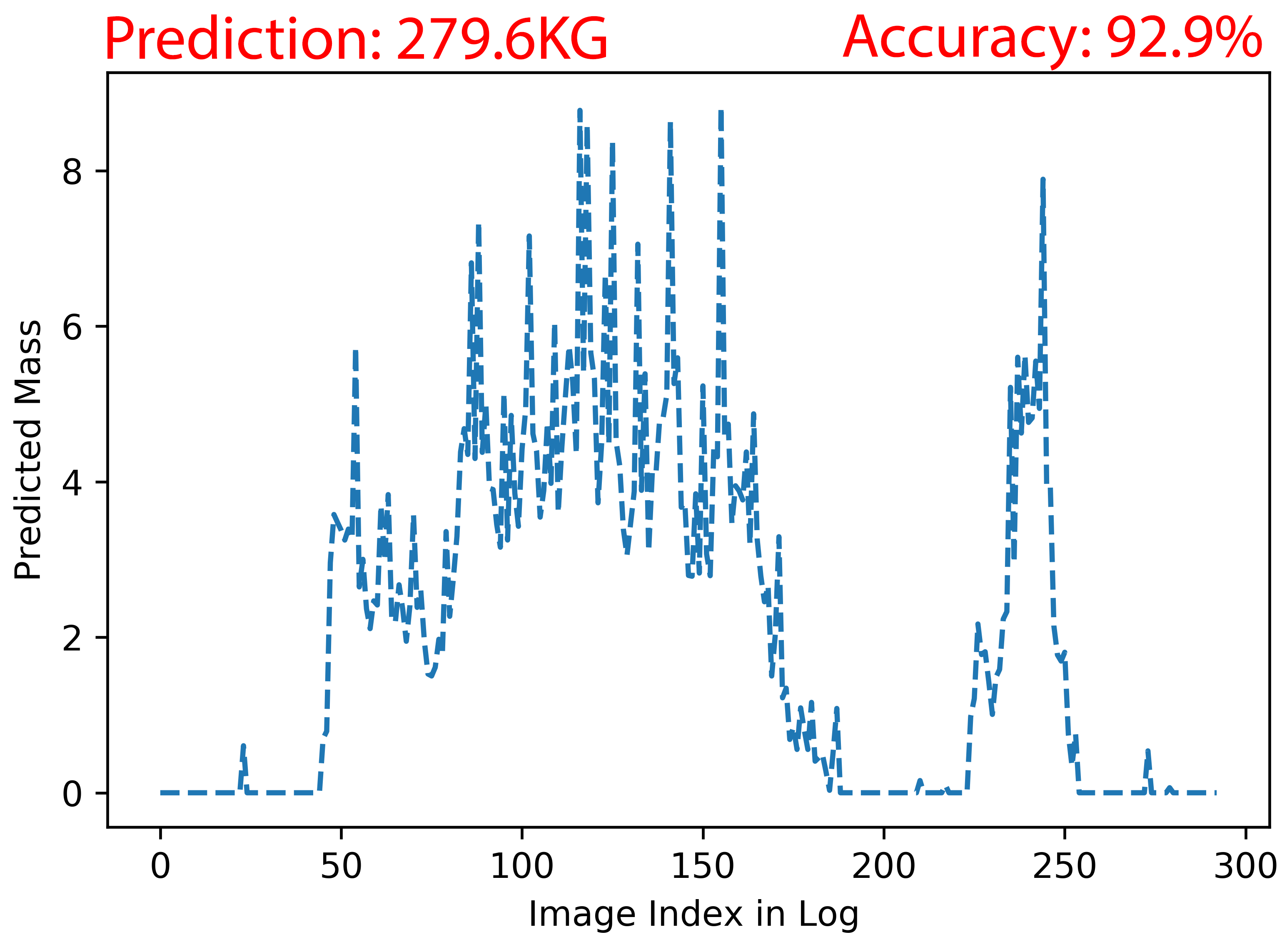}
		\caption{Volume-based signal}
		\label{fig:volume_sig}
	\end{subfigure}
	\hfill
	\begin{subfigure}[b]{0.325\textwidth}
		\centering
		\includegraphics[width=\textwidth]{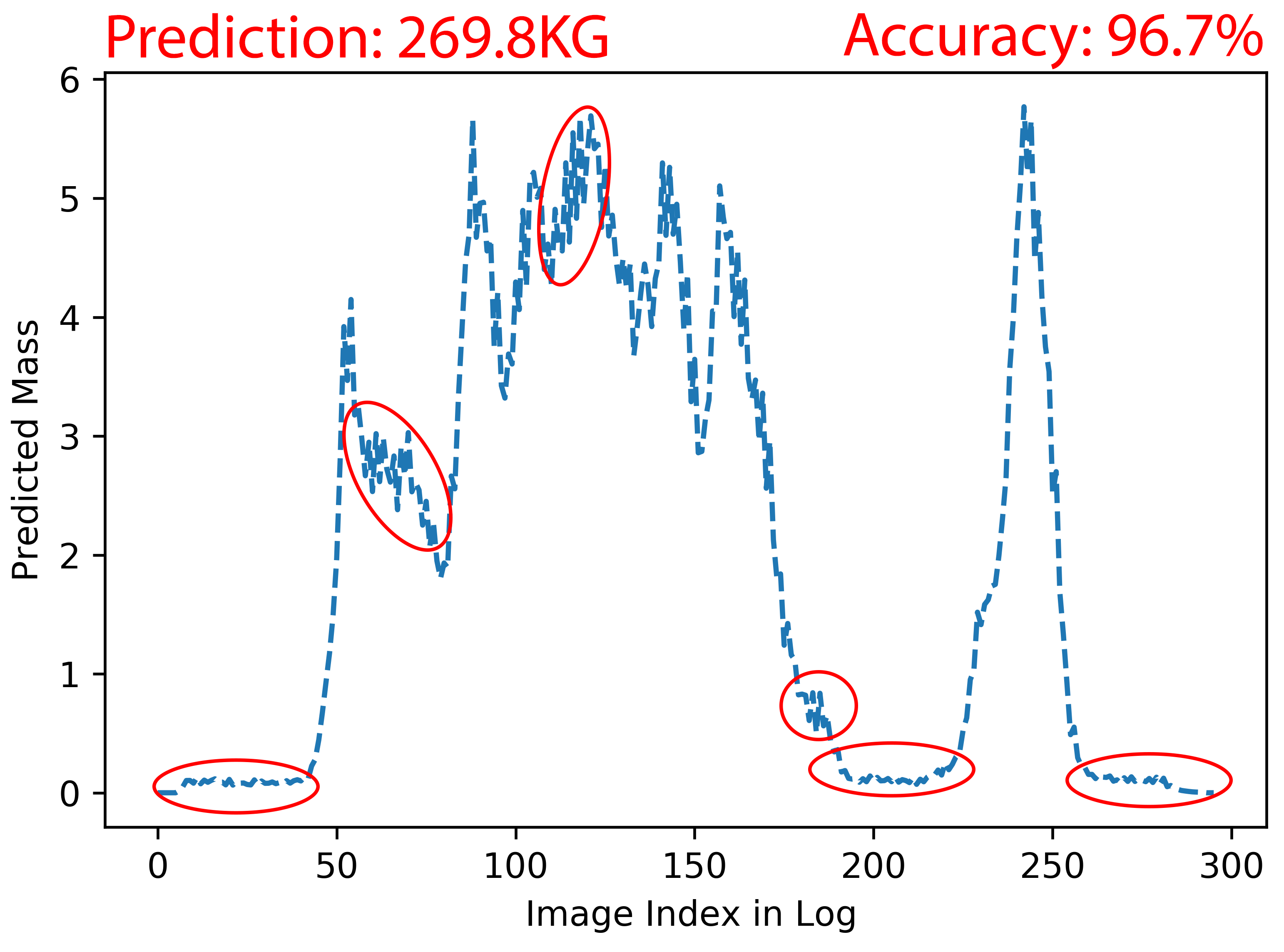}
		\caption{Without smoothness signal}
		\label{fig:elu}
	\end{subfigure}
	\hfill
	\begin{subfigure}[b]{0.325\textwidth}
		\centering
		\includegraphics[width=\textwidth]{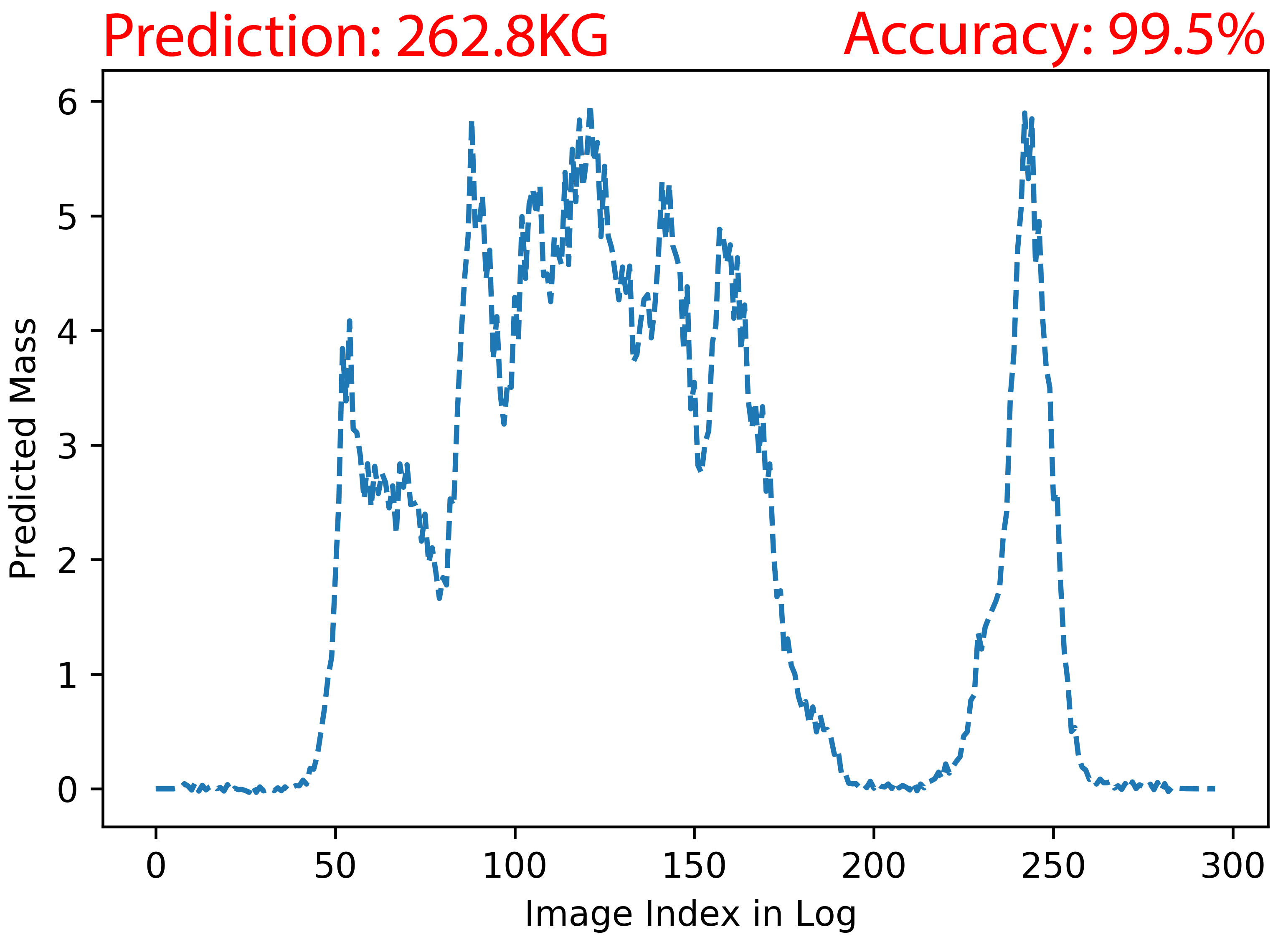}
		\caption{With smoothness signal}
		\label{fig:elu_temp}
	\end{subfigure}
	\caption{Left: Predicted signal of RunX based on the volume algorithm. Middle: Predicted signal of RunX based on DNN algorithm without using temporal smoothness. Right: Predicted signal of RunX based on the DNN algorithm with using temporal smoothness}
	\label{fig:ELURELU_TEMP}
	\vspace{-4mm}
\end{figure}

Training with temporal smoothness for the different developed architectures increased the overall accuracy by 0.6\% on average. Table~\ref{tabl:networks_accuracy} lists the accuracies obtained for RES-9E, RES-16E, and RES-9ER with and without temporal smoothness. Improvements in the signal shape are observed as pointed out in Figure~\ref{fig:elu_temp} when compared to Figure~\ref{fig:elu}, especially the circled regions in red.
\begin{wraptable}{r}{6cm}
	\vspace{-2mm}
	\caption{Summary of architectures accuracies}
	\label{tabl:networks_accuracy}
	\vspace{-4mm}
	\begin{center}
		\begin{tabular}{|p{1.2cm}|p{1.6cm}|p{1.6cm}|}\hline
			%
			&\multicolumn{2}{c|}{\scriptsize \textbf{Test-Set Accuracy} }     \\
			\cline{2-3}
			\textbf{\scriptsize Architecture}   &  \textbf{\centering{\scriptsize W/O Temporal Smoothness}} &  \textbf{\scriptsize W/ Temporal Smoothness}  \\
			\hline
			\textbf{\scriptsize Res-16E} &  \texttt{\scriptsize 94.8\%} &  \texttt{\scriptsize 95.3\%} \\
			\textbf{\scriptsize Res-9E}  &  \texttt{\scriptsize 94.6\%} &  \texttt{\scriptsize 94.9\%}  \\
			\textbf{\scriptsize Res-9ER} &  \textbf{\scriptsize 94.5\%} \tiny($\pm$0.18) &  \textbf{\scriptsize 95.5\%} \tiny($\pm$0.07)  \\
			\hline
		\end{tabular}
		\vspace{-6mm}
	\end{center}
\end{wraptable} Thus far, even without temporal smoothing the signal had very good stability which is much better than the volume-based signal. Figure~\ref{fig:ELURELU_TEMP} shows the predict signal under different scenarios of a run (RunX) trained with RES-9ER with a ground truth of \textbf{261.3KG}. Using temporal smoothness resulted in a more smoothed signal and improved accuracy by 6.6\% and 2.8\% compared to the volume-based and ELU only signals respectively.

\subsection{Comparison to Volume-Based Predictions}
\begin{figure}[!ht]
	\centering
	\vspace{-2mm}
	\begin{subfigure}[b]{0.325\textwidth}
		\centering
		\includegraphics[width=\textwidth]{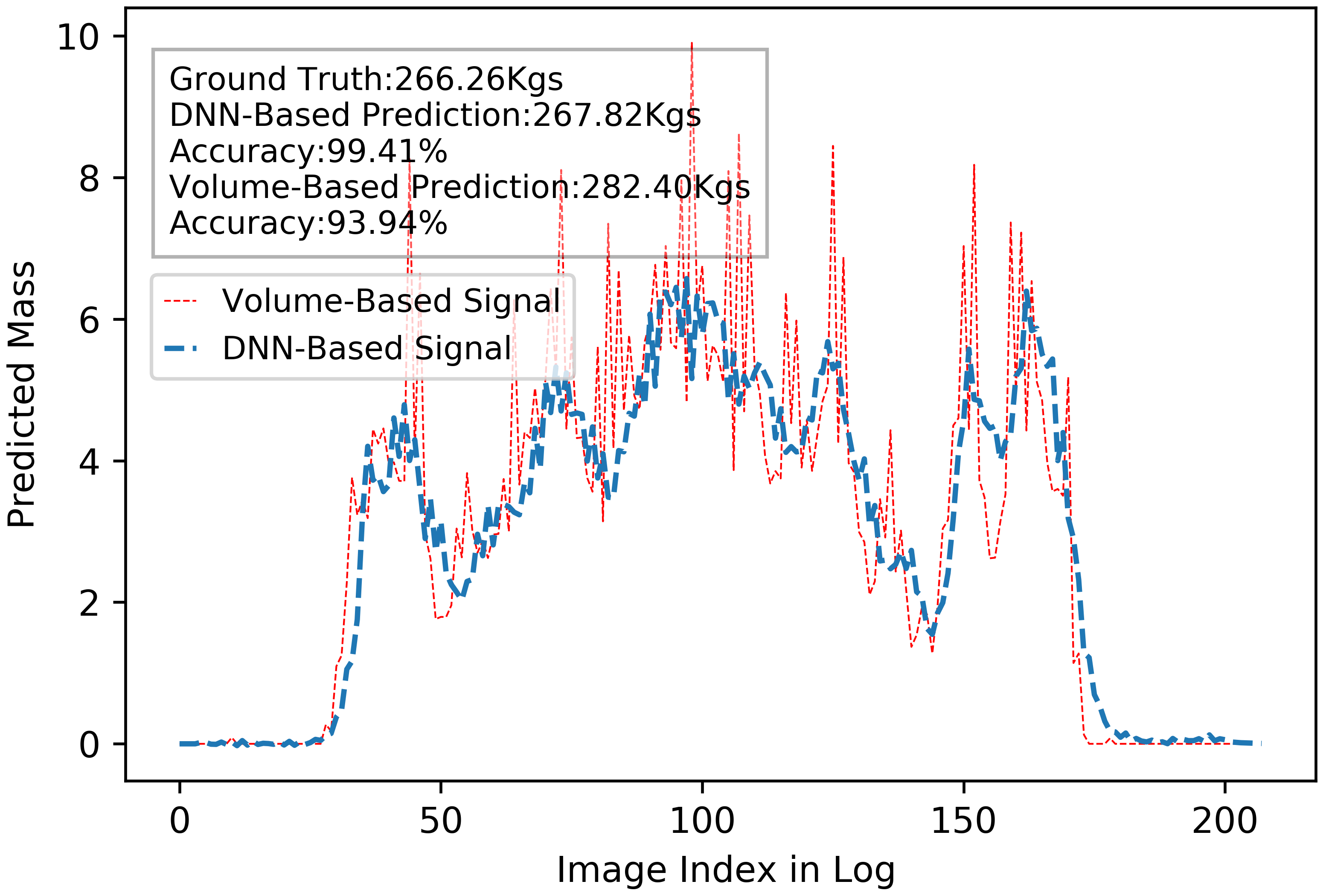}
		\caption{Inc/dec-remental mass flow}
		\label{fig:increment}
	\end{subfigure}
	\hfill
	\begin{subfigure}[b]{0.325\textwidth}
		\centering
		\includegraphics[width=\textwidth]{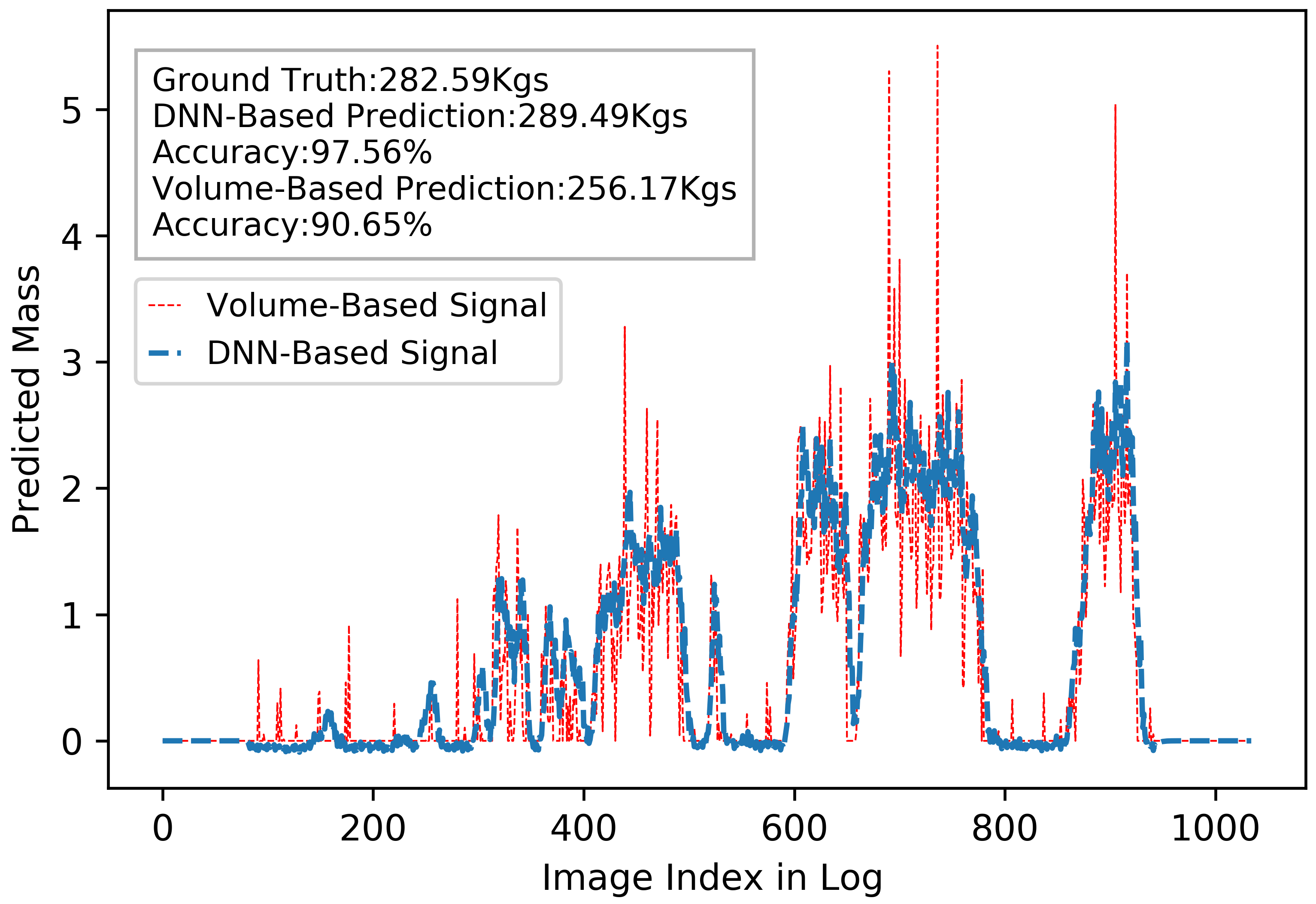}
		\caption{Intermittent mass flow}
		\label{fig:intermittent}
	\end{subfigure}
	\hfill
	\begin{subfigure}[b]{0.325\textwidth}
		\centering
		\includegraphics[width=\textwidth]{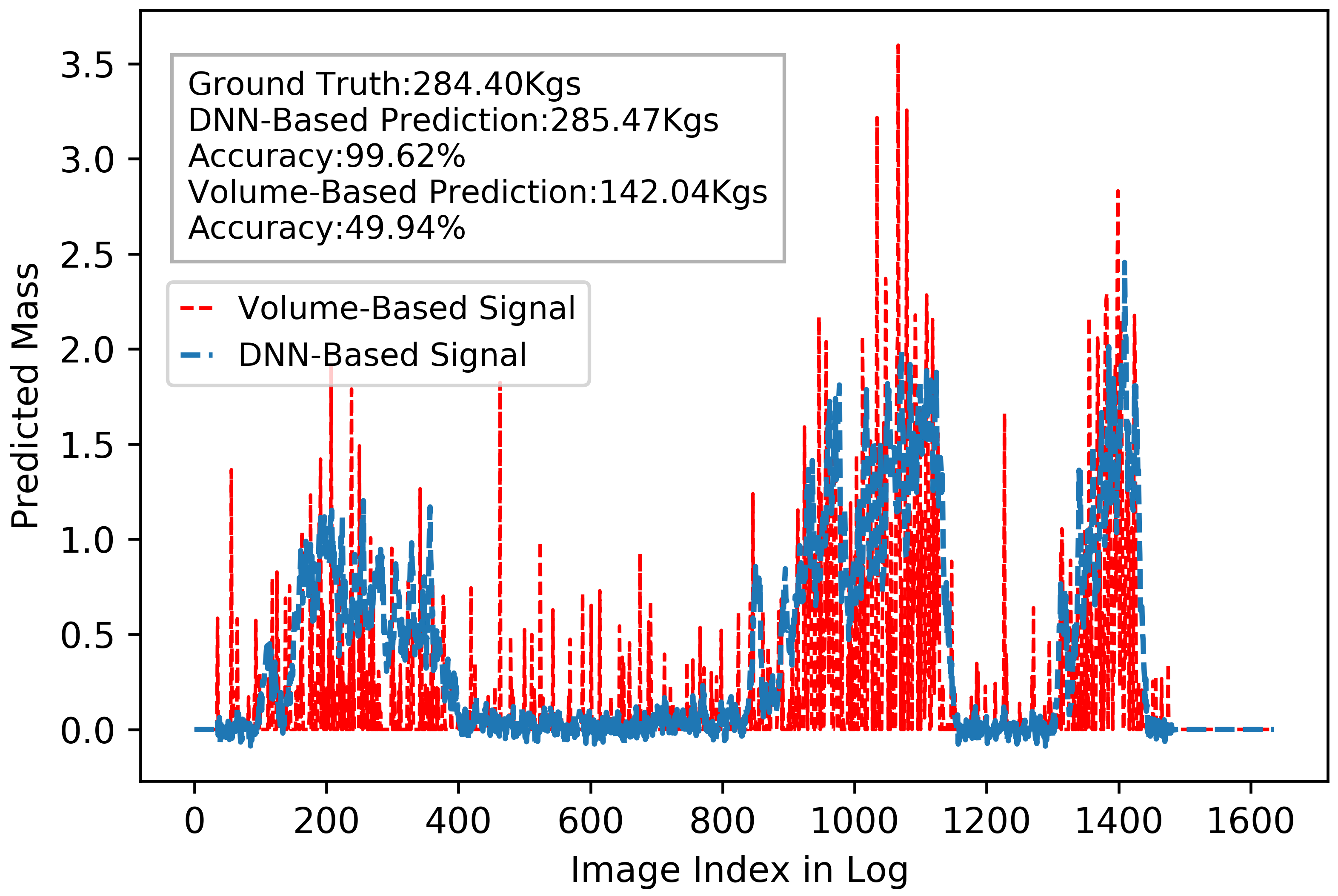}
		\caption{Mass flow with poor lighting}
		\label{fig:poor_light}
	\end{subfigure}
	\caption{Showing the superior performance in terms of stability and accuracy of using a DNN-based image prediction over volume-based prediction of mass}
	\label{fig:variable_conditions}
\end{figure}

We compare the mass predicted signal using Res-9ER against volume-based mass predicted signal using stereo camera in the following scenarios: 1) Incremental and decremental variable flow - Figure~\ref{fig:increment}, 2) Intermittent flow - Figure~\ref{fig:intermittent}, and 3) Poor lighting conditions - Figure~\ref{fig:poor_light}. Further, Figure~\ref{fig:variable_conditions} shows overlays of predicted signals for a number of runs that exhibit the stated characteristics. The DNN-based prediction outperforms the volume-based prediction in each of the scenarios which shows the robustness of the algorithm when posed to variable environmental conditions as opposed to the volume-based signal. We would like to note that the run selected for the poor lighting conditions is one of the runs identified to have too low lighting to use volume estimates on.

\subsection{Error Distribution and Outliers}
\begin{figure}[!ht]
	\centering
	\vspace{-2mm}
	\begin{subfigure}[t]{0.43\textwidth}
		\centering
		\includegraphics[width=\textwidth]{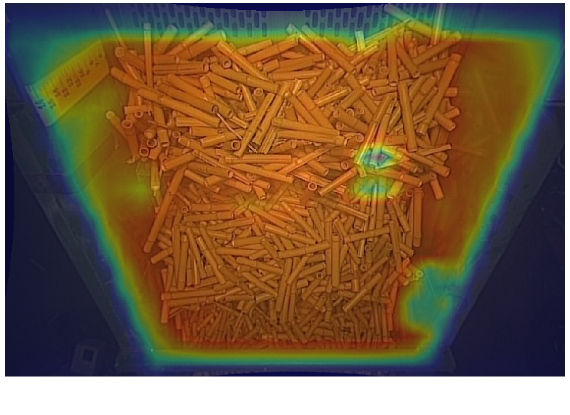}
		\caption{Gradcam of an image from the outlying run}
		\label{fig:outlier_cam}
	\end{subfigure}
	\hfill
	\begin{subfigure}[t]{0.46\textwidth}
		\centering
		\includegraphics[width=\textwidth]{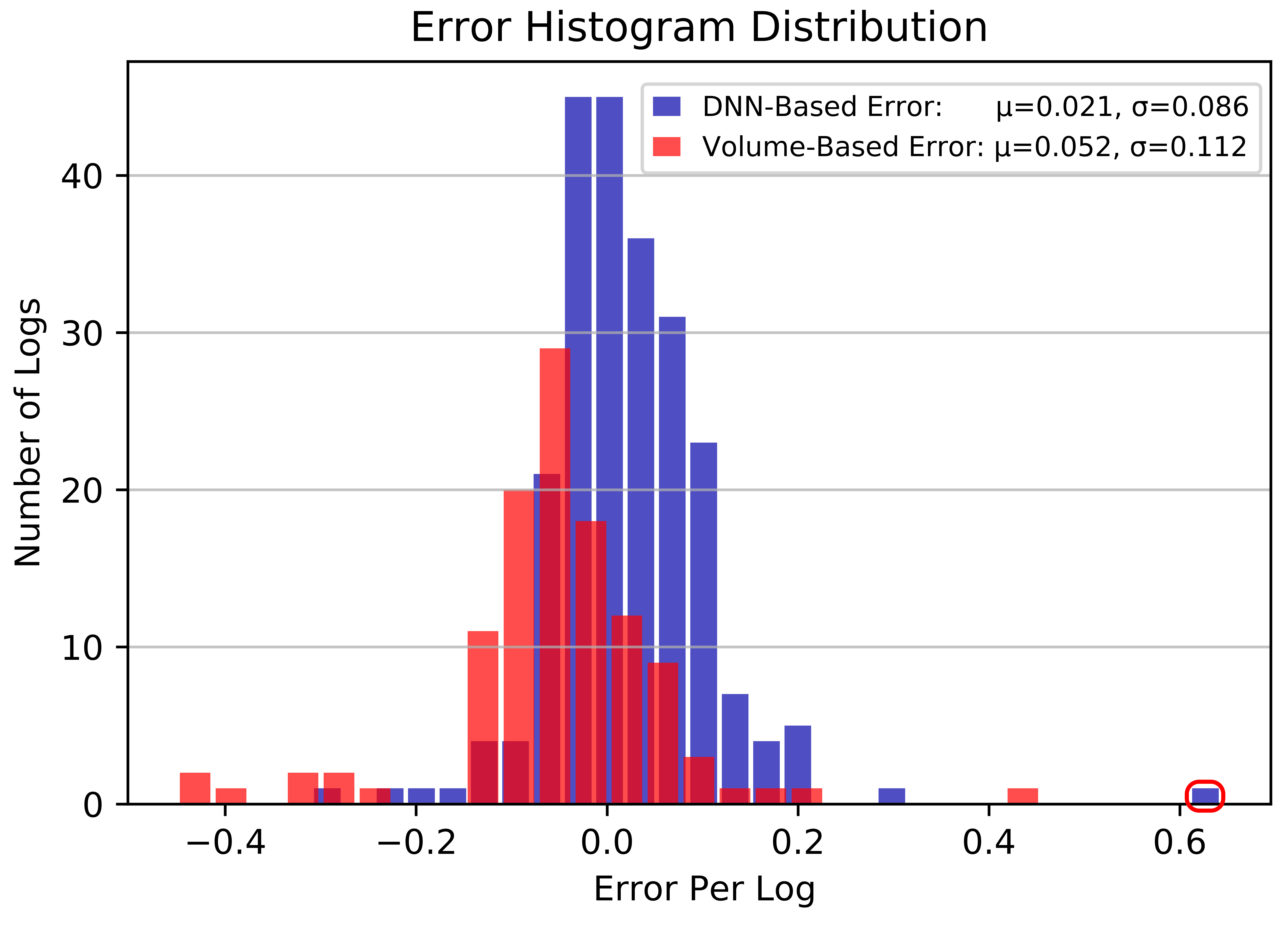}
		\caption{Error distribution of DNN\&Volume predictions}
		\label{fig:dis_err}
	\end{subfigure}
	\caption{Left: Gradcam visualization of an image from the outlying run. Right: Error distribution of predictions by the DNN and the volume algorithms. \textbf{Note}, runs with poor lighting are excluded for the volume algorithm.}
	\label{fig:histograms_outlier}
	\vspace{-4mm}
\end{figure}

Studying error distribution of the dataset helped us understand where the DNN algorithm made poor predictions as well as how to improve generalization. No extreme outliers were found in the validation or test sets, but the training set contained a single outlying run circled in red in Figure~\ref{fig:dis_err}. The total mass prediction of the outlying run was too low ($\sim40\%$ of what it should have been), but yet the DNN mass prediction per image before scaling with elevator speeds and capture time was very high, which was also confirmed by a grad-cam analysis as seen in Figure~\ref{fig:outlier_cam}, where even the sidewalls are being incorrectly incorporated into the prediction of mass. Therefore, we investigated the speed signal and it was found that the outlying run had a very low elevator speed (average of 0.1 m/s) with large variations in speed. It is likely that the optimization process tried to use the reflections of bamboo on the side of elevator to compensate for the low total predicted mass (which was actually a problem of low fidelity from the speed sensor at low elevator speeds). Since this outlier was an extreme corner condition of very low elevator speed that is not at all likely in real applications, thus it could be removed and the network retrained to likely achieve better overall results. Similarly, a corrective action could entail using a speed sensor with higher resolution and fidelity over a broad speed range. Figure~\ref{fig:dis_err} shows histograms of error distribution from the DNN and volume based predictions across the entire dataset. The DNN-based method can be seen to perform much better even though we included a large outlier due to elevator speed abnormalities, and also gave the volume estimates an unfair advantage by excluding low-light runs for which it faired very poorly.

\subsection{Conclusion}
In this work we proposed a semi-supervised algorithm that makes inference on mass flow of material from images by training a DNN with sparse ground truth, and showed improvements over older and more expensive methods that must first acquire a volumetric estimate of the material and then calibrate the density. Future opportunities are considerable. The same methods can be readily applied in other applications where sparse ground truth with some basic assumptions on aggregating the values is possible. A natural extension of this work is to predict mass flow of field harvested sugarcane and grain crops as measured on machine, which experiences many other complicating factors affecting density and visual appearance of the material. Likewise, applying unsupervised learning methods to train disentangled representations of features seems highly feasible and a logical next step.

\bibliographystyle{unsrt}
\bibliography{References}

\begin{thebibliography}{10}

\bibitem{lison2015introduction}
Pierre Lison.
\newblock {\em An introduction to machine learning}.
\newblock Springer: Berlin, Germany, 2015.

\bibitem{marsland2011machine}
Stephen Marsland.
\newblock {\em Machine learning: an algorithmic perspective}.
\newblock Chapman, 2011.

\bibitem{tan2011semi}
Ben Tan, Junping Zhang, and Liang Wang.
\newblock Semi-supervised elastic net for pedestrian counting.
\newblock {\em Pattern Recognition}, 44(10):2297--2304, 2011.

\bibitem{perez2015semi}
Maria P{\'e}rez-Ortiz, JM~Pe{\~n}a, Pedro~Antonio Guti{\'e}rrez, Jorge
  Torres-S{\'a}nchez, C{\'e}sar Herv{\'a}s-Mart{\'\i}nez, and Francisca
  L{\'o}pez-Granados.
\newblock A semi-supervised system for weed mapping in sunflower crops using
  unmanned aerial vehicles and a crop row detection method.
\newblock {\em Applied Soft Computing}, 37:533--544, 2015.

\bibitem{kuznietsov2017semi}
Yevhen Kuznietsov, Jorg Stuckler, and Bastian Leibe.
\newblock Semi-supervised deep learning for monocular depth map prediction.
\newblock In {\em Proceedings of the IEEE Conference on Computer Vision and
  Pattern Recognition}, pages 6647--6655, 2017.

\bibitem{misra2015watch}
Ishan Misra, Abhinav Shrivastava, and Martial Hebert.
\newblock Watch and learn: Semi-supervised learning for object detectors from
  video.
\newblock In {\em Proceedings of the IEEE Conference on Computer Vision and
  Pattern Recognition}, pages 3593--3602, 2015.

\bibitem{jadhav2014volumetric}
U~Jadhav, LR~Khot, R~Ehsani, V~Jagdale, and JK~Schueller.
\newblock Volumetric mass flow sensor for citrus mechanical harvesting
  machines.
\newblock {\em Computers and electronics in agriculture}, pages 93--101, 1993.

\bibitem{shin2012mass}
JUNSU Shin.
\newblock {\em Mass and Size Estimation of citrus Fruit by Machine vision and
  citrus greening diseased fruit detection using spectral analysis}.
\newblock PhD thesis, University of Florida, 2012.

\bibitem{vayrynen2013mass}
Teemu V{\"a}yrynen, Pekka It{\"a}vuo, Matti Vilkko, Antti Jaatinen, and Mika
  Peltonen.
\newblock Mass-flow estimation in mineral-processing applications.
\newblock {\em IFAC Proceedings}, 46(16):271--276, 2013.

\bibitem{zhu2009introduction}
Xiaojin Zhu and Andrew~B Goldberg.
\newblock Introduction to semi-supervised learning.
\newblock {\em Synthesis lectures on artificial intelligence and machine
  learning}, 3(1):1--130, 2009.

\bibitem{sutton2000policy}
Richard~S Sutton, David~A McAllester, Satinder~P Singh, and Yishay Mansour.
\newblock Policy gradient methods for reinforcement learning with function
  approximation.
\newblock In {\em Advances in neural information processing systems}, pages
  1057--1063, 2000.

\bibitem{konstantinovic2007soil}
Miodrag Konstantinovic et~al.
\newblock {\em In-Soil Measuring of Sugar Beet Yield Using UWB Radar Sensor
  System}.
\newblock PhD thesis, Universit{\"a}ts-und Landesbibliothek Bonn, 2007.

\bibitem{liu2018evolutions}
Lufeng Liu, Ye~Yuan, Wei Deng, and Shuixiang Li.
\newblock Evolutions of packing properties of perfect cylinders under
  densification and crystallization.
\newblock {\em The Journal of chemical physics}, 149(10):104--503, 2018.

\bibitem{zhang2006experimental}
Wenli Zhang.
\newblock {\em Experimental and computational analysis of random cylinder
  packings with applications}.
\newblock PhD thesis, Louisiana State University, 2006.

\bibitem{bahdanau2014neural}
Dzmitry Bahdanau, Kyunghyun Cho, and Yoshua Bengio.
\newblock Neural machine translation by jointly learning to align and
  translate.
\newblock {\em arXiv preprint arXiv:1409.0473}, 2014.

\bibitem{he2016identity}
Kaiming He, Xiangyu Zhang, Shaoqing Ren, and Jian Sun.
\newblock Identity mappings in deep residual networks.
\newblock In {\em European conference on computer vision}, pages 630--645.
  Springer, 2016.

\bibitem{resemblejs}
James Cryer and the Huddle~development team.
\newblock Image analysis and comparison.
\newblock \url{https://github.com/rsmbl/Resemble.js}, 2018.
\newblock Accessed: 2019-04-28.

\bibitem{gradcam2017}
Aditya Chattopadhay, Anirban Sarkar, Prantik Howlader, and Vineeth~N
  Balasubramanian.
\newblock Grad-cam++: Generalized gradient-based visual explanations for deep
  convolutional networks.
\newblock In {\em 2018 IEEE Winter Conference on Applications of CV (WACV)},
  pages 839--847. IEEE, 2018.

\bibitem{eager}
Tensorflow team.
\newblock Eager execution in tensorflow.
\newblock \url{https://www.tensorflow.org/guide/eager}, 2019.
\newblock Accessed: 2019-74-21.

\end{thebibliography}
\end{document}